\documentclass[letterpaper, 10 pt, journal, twoside]{IEEEtran}

\usepackage{graphics} 
\usepackage{graphicx}
\usepackage{amsmath} 
\usepackage{amssymb}  
\usepackage{xcolor}
\usepackage{subfigure}
\usepackage{siunitx}
\usepackage{threeparttable}
\makeatletter
\let\NAT@parse\undefined
\makeatother
\usepackage[colorlinks,
            linkcolor=blue,
            anchorcolor=blue,
            citecolor=blue]{hyperref}
\usepackage{algorithm}
\usepackage{algorithmicx}
\usepackage{tabularx,booktabs}
\usepackage{multirow}
\usepackage{CJKutf8}
\usepackage{soul}
\usepackage{tcolorbox}

\usepackage{xspace}
\usepackage{multirow} 
\usepackage{color}
\usepackage{makecell}
\usepackage{graphicx}
\usepackage{float}
\usepackage{colortbl}
\usepackage{bbding}
\usepackage{CJKutf8}
\usepackage{colortbl}
\usepackage{xcolor}
\usepackage{orcidlink}

\definecolor{gray_loc}{RGB}{118,113,113}
\definecolor{bule_lane}{RGB}{91,155,213}
\definecolor{green_rule}{RGB}{116,171,79}
\definecolor{orange_plan}{RGB}{244,173,124}
\definecolor{yellow_drive}{RGB}{255,192,0}

\definecolor{mygray}{gray}{.9}
\definecolor{pastelyellow}{rgb}{1.0, 0.902, 0.59}
\definecolor{myyellow}{rgb}{1.0, 0.988, 0.9}
\newcommand{\systemname}{{PHP}\xspace}

\definecolor{mod_color}{rgb}{0.0, 0.0, 0.0} 

\title{Perception Helps Planning: Facilitating Multi-Stage Lane-Level Integration via Double-Edge Structures}

\begin{document}

\markboth{IEEE Robotics and Automation Letters. Preprint Version. Accepted December, 2024}
{You \MakeLowercase{\textit{et al.}}: Perception Helps Planning: Facilitating Multi-Stage Lane-Level Integration via Double-Edge Structures} 


\author{Guoliang You$^{1}$, Xiaomeng Chu$^{1}$, Yifan Duan$^{1}$, Wenyu Zhang$^{1}$, Xingchen Li$^{1}$, Sha Zhang$^{1}$, Yao Li$^{1}$, \\ Jianmin Ji$^{1,2}$, Yanyong Zhang$^{1,2}$, \emph{Fellow, IEEE}
\thanks{Manuscript received 16 July 2024; revised 3 October 2024; accepted 2 December 2024. This letter was recommended for publication by Editor Hyungpil Moon upon evaluation of the Associate Editor and Reviewers' comments. This work was supported in part by the National Natural Science Foundation of China under Grant 62332016, in part by the National Key R\&D Program of China under Grant 2023YFB4704500, in part by Guangdong Province R\&D Program under Grant 2020B0909050001, and in part by Hunan Province Major Scientific and Technological Program under Grant 2024QK2001.
\emph{(Corresponding author: Jianmin Ji and Yanyong Zhang)}} 
\thanks{$^{1}$Guoliang You, Xiaomeng Chu, Yifan Duan, Wenyu Zhang, Xingchen Li, Sha Zhang, Yao Li, Jianmin Ji and Yanyong Zhang are with the School of Computer Science and Technology, University of Science and Technology of China (USTC), Hefei 230052, China.
{e-mail: \{glyou, cxmeng, dyf0202, wenyuz, starlet, zhsh1, zkdly\}@mail.ustc.edu.cn,  \{jianmin, yanyongz\}@ustc.edu.cn}.}
\thanks{$^{2}$Jianmin Ji and Yanyong Zhang are also with the Institute of Artificial Intelligence, Hefei Comprehensive National Science Center, Hefei 230039.}%
\thanks{Digital Object Identifier (DOI): see top of this page.}
}

\maketitle

\begin{abstract}
\label{sec:abstract}
When planning for autonomous driving, it is crucial to consider essential traffic elements such as lanes, intersections, traffic regulations, and dynamic agents. However, they are often overlooked by the traditional end-to-end planning methods, likely leading to inefficiencies and non-compliance with traffic regulations.
In this work, we endeavor to integrate the perception of these elements into the planning task.
To this end, we propose \textbf{P}erception \textbf{H}elps \textbf{P}lanning (\systemname), a novel framework that reconciles lane-level planning with perception. This integration ensures that planning is inherently aligned with traffic constraints, facilitating safe and efficient driving.
\textcolor{mod_color}{Specifically, \systemname focuses on both edges of a lane for planning and perception purposes, taking into account the positions of both lane edges in Bird's Eye View (BEV), along with attributes related to lane intersections, lane directions, and lane occupancy.}
In the algorithmic design, the process begins with the transformer encoding multi-camera images to extract the above features and predict lane-level perception results. Next, the hierarchical feature early fusion module refines the features for predicting planning attributes. 
\textcolor{mod_color}{Finally, a specific interpreter utilizes a late-fusion process designed to integrate lane-level perception and planning information, culminating in generating vehicle control signals. }
Experiments on three Carla benchmarks show significant improvements in driving scores of 27.20\%, 33.47\%, and 15.54\% over existing algorithms, respectively, achieving state-of-the-art performance, with the system operating up to 22.57 FPS.
\end{abstract}


\section{Introduction}
\label{sec:introduction}
\begin{figure}[t]
  \centering
  \includegraphics[width=\columnwidth]{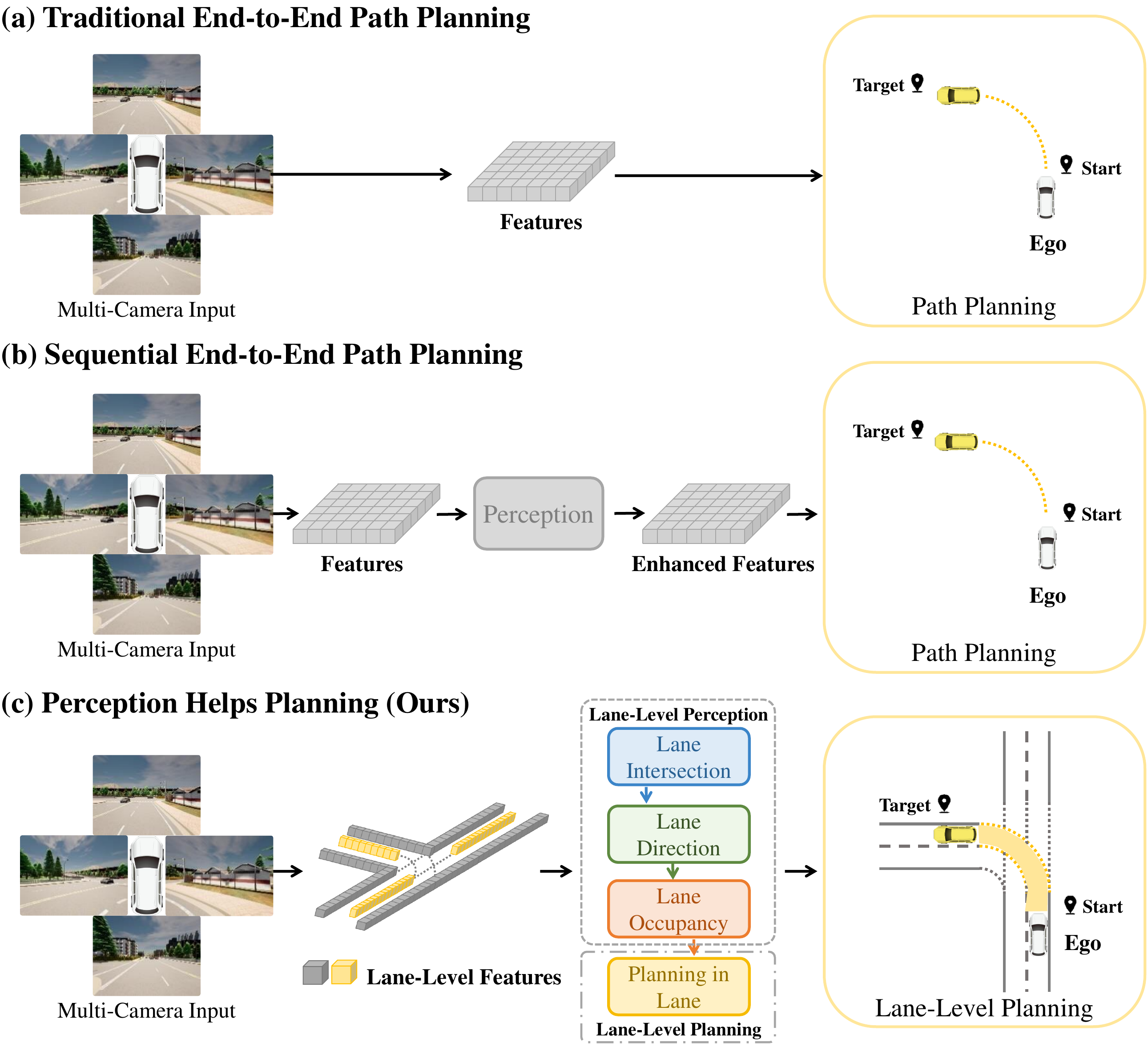}
  \caption{ Comparison of Autonomous Driving Framework: a) Traditional end-to-end framework prioritizes planning policy optimization without considering perception. b) Sequential integration framework enhances planning by incorporating perception into traditional end-to-end planning but lacks interaction between perception and planning. c) Our Perception Helps Planning (\systemname) framework transforms path planning as a lane-level task, integrating multi-level lane-centric perception at both the feature and result levels.}
  \vspace{-12pt}
  \label{fig:teaser}
\end{figure}

\IEEEPARstart{A}{utonomous} driving~\cite{autonomous-survey} plays a vital role in improving the efficiency and safety of transportation systems, intersecting the fields of computer vision and robotics~\cite{perception-3D-survey1,calib-survey, yu2024ldp}. A key challenge lies in devising safe and efficient path planning~\cite{planning-survey}, which must account for various traffic elements such as lane restrictions, intersections, and directions, along with the occupancy of an agent. 
Planning algorithms in autonomous driving can be broadly categorized into two types: rule-based~\cite{perception-3d-BEVFormer,lane-bev-laneDet, rule-planning-rrt, rule-planning-RRV} and end-to-end learning-based planning~\cite{e2e-planning-survey, e2e-nvidia,e2e-pathrl,m2m-stp3, m2m-uniad-Goal-oriented,m2m-neat}.
Traditional rule-based planning uses complex algorithms such as BEV-LaneDet\cite{lane-bev-laneDet} for lane detection and BevFormer\cite{perception-3d-BEVFormer} to predict the positions of vehicles and pedestrians, followed by optimization techniques like RRT\cite{rule-planning-rrt} to find safe paths, with high computational cost.
On the other hand, end-to-end planning, demonstrated by NVIDIA\cite{e2e-nvidia}, maps raw pixels to steering via CNNs, cutting computational needs. However, its efficiency comes at the cost of limited perception of complex environments, affecting planning precision.
To address the perception limitations in traditional models, a variant approach exemplified by the ST-P3 algorithm~\cite{m2m-stp3} implements a sequential strategy. It enhances Bird's Eye View (BEV) representations from multi-cameras using spatiotemporal modules, subsequently utilizing these representations in planning modules.
Despite this, such algorithms inadequately leverage the potential synergy between perception and planning. This leads to a fragmented integration of the two processes, constraining performance.

Given these challenges, the question arises: is it feasible to develop an algorithm that smoothly integrates perception and planning? 
In response, we propose \systemname, a framework where the double-edge data structure facilitates the transformation from path planning to lane-level planning, integrating lane-level perception tasks.
\textcolor{mod_color}{Additionally, to predict elements within this structure, we designed a transformer-based method that utilizes early fusion at the feature level and late fusion at the result level. This approach strengthens the collaboration between perception and planning within this structure.}

\textcolor{mod_color}{Specifically, the data structure encapsulates five components of traffic elements: the position of BEV for each lane edge and attributes for lane intersection, lane direction, lane occupancy, and planning. 
By abstracting traffic information into these components, the data structure effectively captures the traffic scenario's dynamic and static properties.
The algorithm first encodes features from multiple cameras and then employs a specialized transformer to achieve feature embedding for each component. It then utilizes the position of BEV, intersection, direction, and occupancy branches to predict these components.
Subsequently, based on these feature embeddings, a hierarchical feature early fusion module enhances the relevance of features for planning.  In addition, to improve the guidance ability of the target point, we vectorize the target point, guiding the prediction of the planning attribute. 
Finally, an interpreter decodes the data structure, converting and fusion perception and planning at the result level into vehicle control signals.}

This work demonstrates the feasibility of planning at the lane level. It achieves synchronous lane-level perception and planning by deeply integrating both aspects. This integration ensures that planning strategies adhere to traffic regulations and adapt to real-time changes in perception.
In the Carla~\cite{carla} benchmark, extensive tests on Perception Helps Planning (\systemname) confirmed its ability to ensure safety across traffic scenarios and achieve state-of-the-art performance.

In summary, our main contributions are as follows:
\begin{itemize}
\item We introduce the \systemname, a novel lane-level planning framework that transforms path planning into lane-level planning. \systemname integrates perception with planning and ensures planning compliance with traffic rules.
\item We develop an algorithm that extracts lane-level traffic features from multi-camera images, integrating early-fusion of features level with late-fusion of result-level to significantly enhance safety and efficiency.
\item We conducted experiments on three benchmarks, where \systemname 
 outperformed previous algorithms with driving score improvements of 27.20\%, 33.47\%, and 15.54\%, and up to 22.57 FPS, achieving state-of-the-art performance.
\end{itemize}

\section{Relate work}
\label{sec:relate_work}

\begin{figure*}[t]
  \centering
  \includegraphics[width=\linewidth]{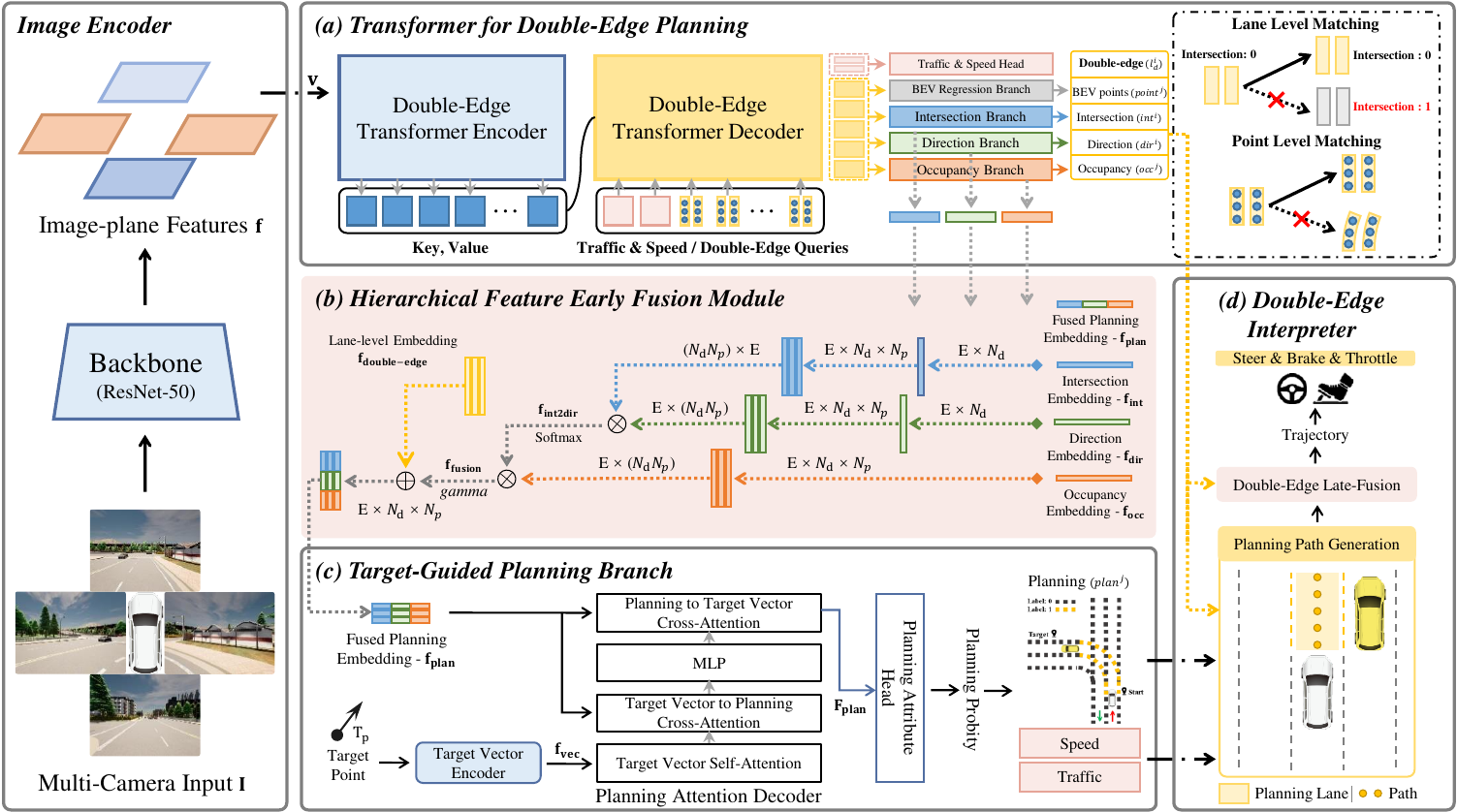}
  \caption{\textcolor{mod_color}{The \systemname begins with an image encoder that uses a ResNet to extract features $\mathbf{f}$ from multi-camera inputs $\mathbf{I}$, which are sequenced with positional encoding to form input $\mathbf{v}$ for the transformer. (a) The transformer processes input $\mathbf{v}$, extracting lane features $\mathbf{f_{double-edge}}$ and predicting BEV points $\mathit{point^{j}}$ . Simultaneously, it extract features of intersection $\mathbf{f_{int}}$, direction $\mathbf{f_{dir}}$, and occupancy $\mathbf{f_{occ}}$ through dedicated branches, and utilizes these predicted attributes $\mathit{int^i}$, $\mathit{dir^i}$, and $\mathit{occ^j}$. (b) The fusion module integrates $\mathbf{f_{int}}$ and $\mathbf{f_{dir}}$ into a probability matrix $\mathbf{f_{int2dir}}$, which, when merged with $\mathbf{f_{occ}}$, forms a comprehensive lane feature $\mathbf{f_{fusion}}$. This feature, combined with $\mathbf{f_{double-edge}}$, generates a planning feature $\mathbf{f_{plan}}$ using a learnable parameter $\mathit{gamma}$. (c) The target-guided planning branch enhances the interaction between the target point $\mathit{T_p}$ and features $\mathbf{f_{plan}}$ through attention mechanisms for predicting planning attributes $\mathit{plan^i}$. (d) Finally, the interpreter fuses and transforms the perception and planning information at the resulting level into a path, incorporating traffic signals and speed to generate a trajectory for control commands. The symbol $\oplus$ represents point-wise addition, while $\otimes$ denotes matrix multiplication.}}
\vspace{-15pt}
  \label{fig:tgd_arc}
\end{figure*}

In this section, we provide an overview of rule-based planning algorithms~\cite{calib-survey,perception-3D-survey1,lane-survey,planning-survey} and end-to-end learning-based planning algorithms in autonomous driving~\cite{e2e-planning-survey}.

\noindent\textbf{Traditional Rule-based Planning Algorithm. }
Traditional rule-based algorithms integrate perception and planning, forming the navigation systems. The perception accurately models the environment. It employs Simultaneous Localization and Mapping (SLAM) algorithms~\cite{slam-pf,add_slam_makeyouslam} to enable autonomous systems to construct and update environmental maps and determine their positions within these maps. \textcolor{mod_color}{Camera-based 3D object detection techniques~\cite{perception-trackingez,perception-3d-oa-bev,perception-3d-neighbor-vote,add_3d_det_vsrd} or multi-sensor fusion 3D object detection techniques~\cite{add_3d_det_sparselif} identify, recognize, and locate objects, enriching the environmental model. Lane detection algorithms~\cite{lane-bev-laneDet,lane-maptr,lanesegnet,add_lane_det_lane2seq,add_lane_det_lanecpp} delineate traffic constraints by detecting lanes, contributing to the environmental model.}
Based on the environmental model provided by the perception, planning employs search-based algorithms such as Rapidly Exploring Random Trees (RRT) and its variants~\cite{rule-planning-rrt, rule-planning-rrt-growth, rule-planning-rrt-connect, rule-planning-rrt-probabilistic}, along with the A Star (A*) search algorithm~\cite{rule-base-a-star}, to plan a safe path to the destination. 
These algorithms take advantage of perception data for safe and efficient path planning.

\noindent\textbf{End-to-End Learning-based Planning Algorithms. }
End-to-end planning aims to simplify the system by using neural networks to directly output planning policies, such as path or steering control signals. The initial methodologies directly predicted these policies~\cite{e2e-cilrs,e2e-lbc,e2e-nvidia,e2e-pathrl,e2e-mit-variational}, sidestepping the environmental perception and understanding, leading to limited performance and interpretability. Such as NVIDIA~\cite{e2e-nvidia} used the camera as an input to train a neural network that directly outputs control signals. LBC~\cite{e2e-lbc} utilized mimicking techniques to train image networks with supervision from a privileged model.
To mitigate these drawbacks, variants of end-to-end methods integrate perception into planning sequentially or as auxiliary tasks~\cite{m2m-wor,m2m-interfuser,m2m-stp3,m2m-lav,m2m-multi-moda-fusion,m2m-neat,m2m-roach,m2m-transfuser,m2m-uniad-Goal-oriented,m2m-vad}. This integration enhances the planning system's environmental comprehension, which improves performance.
In this context, NEAT~\cite{m2m-neat} introduced neural attention fields for infusing reasoning capabilities into end-to-end driving models. 
ST-P3~\cite{m2m-stp3} proposes a spatial-temporal feature learning scheme to generate more representative features for perception, prediction, and planning.
\section{Methodology}
\label{sec:methodology}
\subsection{Overall Framework}
As in Figure \ref{fig:tgd_arc}, \systemname introduces a novel method that integrates perception and planning at the lane level.
At the core of \systemname is the double-edge, which contains the positions of BEV for each lane edge, lane-level attributes (i.e., intersection and direction),  and point-level attributes (i.e., occupancy and planning). 
\textcolor{mod_color}{In the algorithm, the transformer for planning (Figure~\ref{fig:tgd_arc}(a)) designs a double-edge query for feature retrieval. This feature decodes the positions of BEV for each lane edge via a BEV regression branch while extracting intersections, direction, and occupancy features for hierarchical feature fusion and predicting these attributes through the corresponding branches. 
The hierarchical feature early fusion module (Figure~\ref{fig:tgd_arc}(b)) employs attention mechanisms to fuse intersection, direction, and occupancy features, enriching the planning feature with a thorough understanding of traffic scenarios. 
The target-guided planning branch (Figure~\ref{fig:tgd_arc}(c)) uses cross-attention to enhance the dynamic responsiveness to target points, thus improving planning accuracy. 
Lastly, a specific interpreter (Figure~\ref{fig:tgd_arc}(d)) translates the combined perception and planning data into vehicle control signals.}

\begin{figure}[t]
  \centering
  \includegraphics[width=0.9\linewidth]{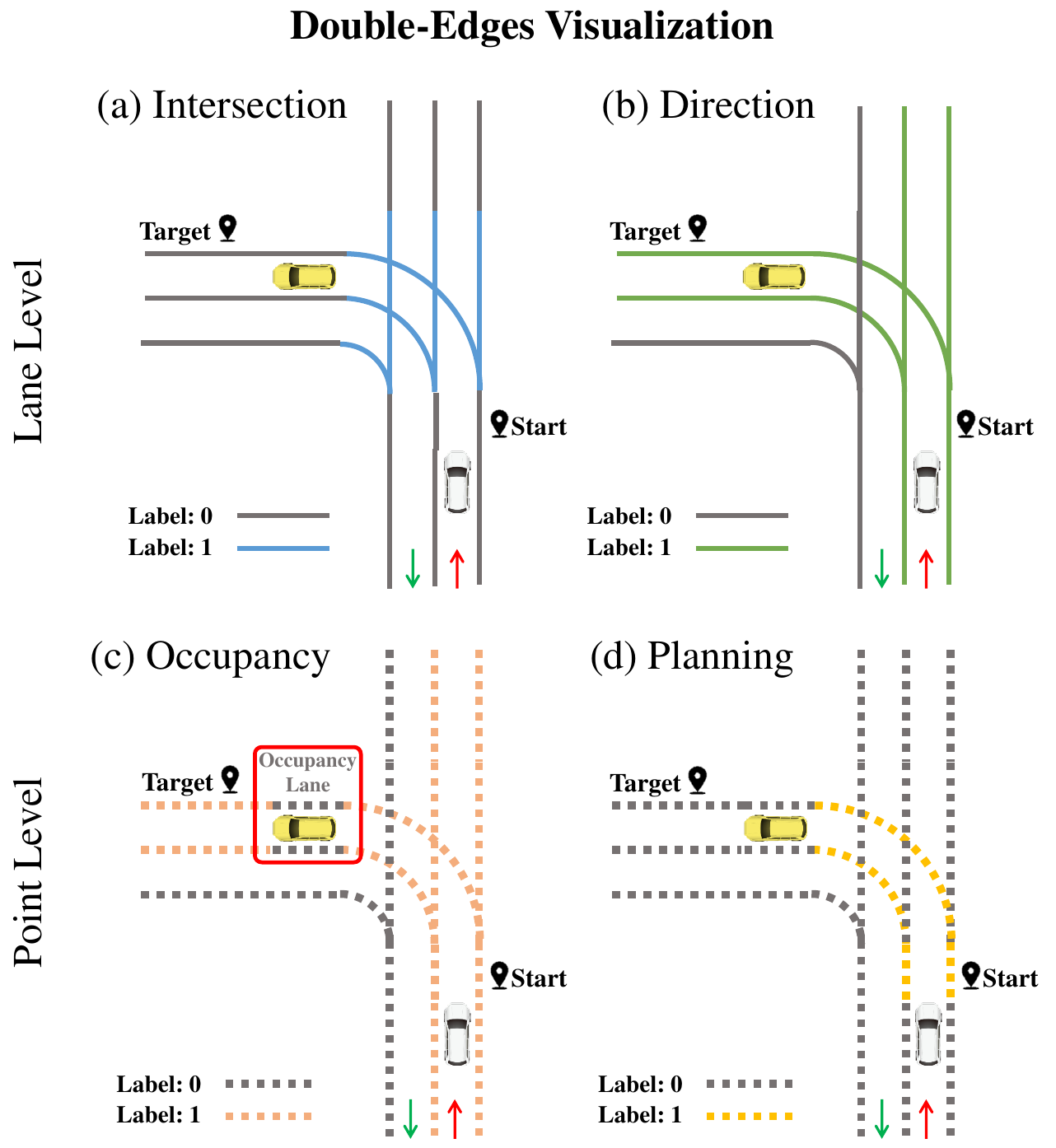}
  \caption{Visualizes double-edge in a traffic scenario ${L}$, with \textcolor{bule_lane}{blue} and \textcolor{green_rule}{green} detailing lane-level attributes for intersections and directions. \textcolor{orange_plan}{orange} and \textcolor{yellow_drive}{yellow} detail point-level attributes, marking unoccupied and planning lanes.}
  \vspace{-15pt}
  \label{fig:dual_lane_data}
\end{figure}

\subsection{Double-Edge Data Structure}
\label{method:dual-lane-data}
We define $N_{d}$ double-edge data ($l^{i}_{d}$) that incorporate lane and point-level traffic information to describe environment $L$:
\textcolor{mod_color}{
\begin{equation}
\begin{aligned}
{ L} &=\left\{{ l}_{{d }}^i\right\}_{i=0}^{\mathrm{N}_d},\\
 { l }_{ {d }}^i &=\left( { edge  }_{ {l }}^i,  { edge  }_{ {r }}^i,  { int }^i,  { dir }^i\right),\\
 { int }^i &=\{0 \ { or\ } 1\},  { dir }^i=\{0 \ { or\ } 1\},
 \end{aligned}
\end{equation}
}
\textcolor{mod_color}{
where $l^{i}_{d}$ includes the lane's left and right $edges_i$, along with its lane-level intersection $int^i$ and direction $dir^i$ attributes.
The variables $int^{i}$ and $dir^{i}$ indicate whether a lane is an intersection and if the lane's direction is aligned with the direction of the ego vehicle's travel. }
$edge^i$ encompasses points in BEV and attributes for corresponding points, including occupancy and planning attributes that indicate the point level. Each $edge$ consists of $\frac{N_p}{2}$ elements, defined as:
\textcolor{mod_color}{
\begin{equation}
\begin{aligned}
 {edge}^{i} &=\left\{ { point }^{j}, { occ }^{j}, { plan }^{j}\right\}_{j=0}^{\frac{N_p}{2}},\\
 {point}^{j} &=\{\mathit{x}, \mathit{y}\}, \operatorname{occ}^{j}=\{0  {\ or\ } 1\},  { plan }^{j}=\{0  {\ or\ } 1\},
\end{aligned}
\end{equation}
}
where $occ^{j}$ and $plan^{j}$ indicate whether the lane is occupied by traffic agents (e.g., pedestrians, vehicles) and whether it is selected for planning. 
In Figure~\ref{fig:dual_lane_data}, $point^{j}$ illustrates the location of the edge using dashed and solid lines to represent lane-level and point-level visualizations, respectively. Gray denotes attribute values of zero, while blue, green, orange, and yellow indicate one value, visualizing different properties.
\textcolor{mod_color}{In Figure~\ref{fig:dual_lane_data}, (a) shows blue identifying lanes as intersections, green in (b) as lanes adhering to ego vehicle travel direction, orange in (c) highlighting points in lanes unoccupied by pedestrian and vehicle, and yellow in (d) as point selected for planning.}
This structure transforms path planning into a lane-level task, allowing concurrent perception and planning.

\subsection{Transformer for Double-Edge Planning}
\label{method:dual-lane-transformer}
To acquire lane-level features that support the learning of lane-level perception and planning tasks, we have incorporated the transformer, applying a double-edge query to learn lane-level features, as depicted in Figure \ref{fig:tgd_arc}(a).

\noindent{\textbf{Double-edge transformer encoder. }}
Initially, for each image input $\mathbf{I} \in \mathbb{R}^{3 \times H_0 \times W_0}$ , we employ a ResNet-50~\cite{tool-ResNet} to extracting features $\mathbf{f} \in \mathbb{R}^{C \times H \times W}$. The values for $C$, $H$, and $W$ are defined as $C=256$, $H = \frac{H_0}{32}$, and $W = \frac{W_0}{32}$. The dimension of the transformer hidden layer is $E$. For each feature ${\mathbf{f}}$, the transformer encoder applies a ${1×1}$ convolution to generate a lower-channel feature $\mathbf{z} \in \mathbb{R}^{E \times H \times W}$. Next, we simplify the spatial dimensions of ${\mathbf{z}}$ into a sequence, forming ${E \times HW}$ tokens. A fixed sinusoidal positional encoding $\mathbf{e} \in \mathbb{R}^{E \times HW}$ is then added to each token to preserve positional information within each sensor input:
\begin{equation}
{\mathbf{v}}_i^{(x, y)}={\mathbf{z}}_i^{(x, y)}+{\mathbf{e}}^{(x, y)},
\end{equation}
where ${\mathbf{z}_i}$ represents the tokens extracted from the ${i-th}$ view, and ${x}$ and ${y}$ denote the token's coordinate index in that sensor.
Finally, we concatenate the tokens from all sensors and pass them through a transformer encoder comprising ${K}$ standard transformer layers. 
Each layer \( K \) consists of Multi-Headed Self-Attention, MLP blocks, and layer normalization.

\begin{figure}[t]
  \centering
  \includegraphics[width=\linewidth]{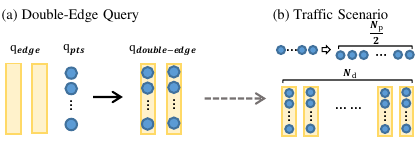}
  \caption{\textcolor{mod_color}{(a) Each query $\mathbf{q_{double-edge}}$ consists of pair of edge query, $\mathbf{q_{edge}}$, each edge query comprising a set of query points, $\mathbf{q_{pts}}$. (b) A scenario features $N_d$ such $\mathbf{q_{edge}}$ pairs, and each lane within a $\mathbf{q_{edge}}$ contains $\frac{N_p}{2}$ query points.}}
  \vspace{-15pt}
  \label{fig:dual_lane_query}
\end{figure}

\noindent{\textbf{Double-edge transformer decoder.} }
\textcolor{mod_color}{The transformer decoder's key function is to query the lane-level features using $K$ layers multi-head self-attention, which predict points in BEV and attributes for intersection, direction, and occupancy.
We have designed multiple queries $\mathbf{q_{double-edge}} \in \mathbb{R}^{E \times N_p}$ to query lane-level features. A total of ${N_{d}}$ such queries are deployed, as shown in Figure~\ref{fig:dual_lane_query}. 
Additionally, we introduce the speed query $\mathbf{q_s} \in \mathbb{R}^{E \times 1}$  and traffic query  $\mathbf{q_t} \in \mathbb{R}^{E \times 1}$ to query the maximum speed `${Speed}$' allowed on the lane selected by planning attribute and to perceive traffic signals, respectively. 
In addition, based on lane-level features, we have devised a BEV regression branch capable of predicting the points in BEV, with size ${N_{d} \times N_{p} \times 2}$. We then processed the lane-level features through three different feature extraction and attribute prediction branches to achieve prediction results for intersection (${N_{d} \times 1}$), direction (${N_{d} \times 1}$), and occupancy (${N_{d} \times N_{p} \times 1}$). Throughout this process, we preserve these features of intersection $\mathbf{f_{int}} \in \mathbb{R}^{E \times N_{d}}$, direction $\mathbf{f_{dir}} \in \mathbb{R}^{E \times N_{d}}$, occupancy $\mathbf{f_{occ}} \in \mathbb{R}^{E \times N_{d} \times N_{p}}$, and lane-level features $\mathbf{f_{double-edge}} \in \mathbb{R}^{E \times N_{d} \times N_{p}}$, laying the foundation for the hierarchical lane fusion module.}

\subsection{Hierarchical Feature Early Fusion Module}
\label{method:dual-lane-fusion}

Figure \ref{fig:tgd_arc}(b) introduces the hierarchical feature early fusion module, which effectively integrates dynamic and static traffic information through attention at lane and point levels, enabling a more detailed understanding of traffic scenes. 
First, we extend intersection and direction embedding across $N_{d}$ dimensions at the lane level, achieving dimensions of $E \times N_{d} \times N_{p}$. This is because each point within a double-edge shares identical intersection and direction attributes. 
Next, we multiply the intersection embedding with the direction embedding to form a feature matrix $\mathbf{f_{int2dir}} \in \mathbb{R}^{N_dN_p \times N_dN_p}$. 
This multiplication integrates intersection and direction attributes at the lane level, defined as:
\textcolor{mod_color}{
\begin{equation}
\begin{aligned}
    \mathbf{f_{ {int2dir }}}=\mathit{R}\left(\mathbf{f_{ {int }}},\left[ {E, }N_d N_p\right]\right)^{\mathrm{T}} 
   \odot \mathit{R}\left(\mathbf{f_{ {dir }}},\left[{ E, } N_d N_p\right]\right),
\end{aligned}
\end{equation}
}
where $\mathit{R}$ denotes reshape operation. Subsequently, We convert the feature matrix $\mathbf{f_{int2dir}}$ into a probability matrix via softmax, quantifying the significance of each feature. 

Then, to integrate point-level occupancy embedding, we obtain the feature matrix $\mathbf{f_{fusion}} \in \mathbb{R}^{E \times N_{d}N_{p}}$ by performing pointwise multiplication between this probability matrix and the occupancy embedding.
This operation serves as attention, highlighting occupancy embedding relevant to an intersection and direction lane, thus focusing on compliant intersections and lane direction areas, $\mathbf{f_{fusion}}$ is defined as:
\textcolor{mod_color}{
\begin{equation}
\mathbf{f_{{fusion }}}=\mathit{R}\left(\mathbf{f_{ {occ }}},\left[ { E }, N_d N_p\right]\right) \odot \operatorname{Softmax}\left(\mathbf{f_{ {int2dir }}}\right).
\end{equation}
}

Finally, by integrating the matrix $\mathbf{f_{fusion}}$ with the lane-level embedding $\mathbf{f_{double-edge}}$, we obtain the fused planning embedding  $\mathbf{f_{plan}} \in \mathbb{R}^{E \times N_{d} \times N_{p}}$, which improves understanding of static and dynamic information for prediction.
This integration employs weighted fusion, controlled by a learnable parameter ${gamma}$, which dictates the fusion degree between the original and new features. $\mathbf{f_{plan}}$ is defined as:
\textcolor{mod_color}{
\begin{equation}
\mathbf{f_{ {plan }}}\!\!=\!\mathit{R}\left(\mathbf{f_{ {fusion }}},\left[{ E },\! \mathrm{N}_d,\!\! \mathrm{~N}_p\right]\right) \times  { gamma }+\mathbf{f_{ {double-edge }}}.
\end{equation}
}

\subsection{Target-Guided Planning Branch}
\label{method:drive-lane-prediction}
We aim to assist autonomous driving systems in selecting the optimal planning lane guided by a specified target point. To accomplish this, we introduce a target vector encoder and a planning attention decoder, as shown in Figure \ref{fig:tgd_arc}(c).
The target vector encoder, $\mathit{Encoder_{vec}}$, encodes the target point, ${T_p}$, into a target vector embedding. It then applies position embeddings from random spatial distributions, transforming spatial coordinates into a higher-dimensional target vector embedding $\mathbf{f_{vec}}$. This enables the model to extract and utilize destination position information more effectively. It can be formulated as:
\textcolor{mod_color}{
\begin{equation}
\mathbf{f_{{vec} }}=\mathit{Encoder}_{{vec} }\left(\mathit{T_p}\right).
\end{equation}
}

The planning attention decoder starts with target vector self-attention ($\mathit{TSA}$) on the encoded target vector embedding $\mathbf{f_{{vec}}}$, followed by target vector to planning cross-attention ($\mathit{TPA}$) on the planning embedding, which is then processed through a multilayer perceptron ($\mathit{MLP}$). Subsequently, planning to target vector cross-attention ($\mathit{PTA}$) amplifies the most relevant planning embedding, $\mathbf{F_{{plan}}}$, to the target vector. It enhances the planning embedding's ability to highlight the relevant features of the target vector. $\mathbf{F_{{plan}}}$ is formulated as:
\textcolor{mod_color}{
\begin{equation}
\!\!\mathbf{F}_{\! {plan }}\!\!=\!\mathit{PTA}\!\left(M L P\!\left(TPA\!\left(TSA\!\left(\mathbf{f_{vec }}\!\right)\!,\!\mathbf{f_{{plan }}}\!\right)\right)\!,\!\mathbf{f_{ {plan }}}\!\right)\!,
\end{equation}
}
where $\mathit{TPA}$ enhances planning features relevant to the target vector through cross-attention, $\mathit{PTA}$ reinforces this relevance with additional cross-attention. These stages effectively capture information relevant to the target points within the planning embedding.
Finally, the planning attribute head processes the amplified embedding $\mathbf{F_{plan}}$ to predict the probability of planning attributes (${N_{d} \times N_{p} \times 1}$). 

\subsection{Double-Edge Interpreter}
\label{method:dual-lane-interpreter}
This interpreter integrates planning path generation and late fusion for safer navigation.

\noindent\textbf{Planning Path Generation. }
Firstly, we use left and right edge points in each double-edge to reconstruct lane information and assign attributes. These attributes include intersection and direction for lane level and occupancy and planning for point level, as shown in Figure \ref{fig:dual_lane_data}.
Secondly, we select edge points marked by a planning attribute value of `1' (indicating suitability for planning) to construct the path:
\begin{equation}
{ Path }=\!\!\bigcup_{\tiny j=1}^{\tiny {N_d}\times{\frac{N_p}{2}}}\!\!\left\{\! \frac{ { point }_{ {l }}^{\mathrm{j}}+{ point }_{ {r }}^{\mathrm{j}}}{2} \bigg| { plan }_{l}^j,{ plan }_{r}^j=1\!\!\right\}\!.\!
\end{equation}

\noindent\textbf{Double-Edge Late-fusion. }
To leverage the perceived information—intersection, direction, and occupancy, we integrate this information using a late-fusion strategy to enhance safety. Initially, direction filters out non-compliant occupancy data, focusing on lanes adhering to traffic regulations. We then assess each occupancy attribute through the planning attribute at the point level. When the planning attribute is `1' and occupancy is `0' (signifying traffic participants within planning lanes), the respective planning path is marked as a stopping path, represented by ${Stop}$.
In addition, if the planning path length is `1', indicating an extremely short path, we classify this path as a stopping path to ensure safety, indicating an immediate need for avoidance action. It can be formulated as: 
\textcolor{mod_color}{
\begin{equation}
    {Stop}=\begin{cases}
    {True},\ & \textit{if\ plan} ^ j=1 , occ ^ j=0\\ 
    {True},  & \textit{if}\  {lenght}({Plan}_{path})=1\\
    {False},  & Otherwise
    \end{cases}.
\end{equation}
}

Finally, the planning path ($Path$), speed ($Speed$), and stop information ($Stop$) are integrated as trajectory inputs to the Model Predictive Control (MPC)\cite{tool-mpc} for generating control commands. With integration, this interpreter ensures safety and enhances interpretability.
\textcolor{mod_color}{
\begin{equation}
    { Trajectory }=\left[ { Path }, { {Speed} }, { {Stop} }\right].
\end{equation}
}

\vspace{-10pt}
\subsection{Loss Function}
\label{method:loss}
Based on these results, our loss function, inspired by MapTR~\cite{lane-maptr}, includes the alignment cost and prediction loss.

\noindent\textbf{Double-Edge Alignment Cost. }
We aim to determine optimal alignment $\hat{\pi}$ between predicted and ground truth data for training. \textcolor{mod_color}{Specifically, \({\pi}(i)\) denotes the predicted element that corresponds to the \(i\)-th ground truth element in the optimal alignment.} This involves aligning the predicted intersection attribute $\hat{v}_i$ with its ground truth $v_i$, and the predicted points in BEV $\hat{p}_i$ with the ground truth $p_i$, defined as:
\begin{equation}
\!\!\!\hat{\pi}\!=\!\arg\!\!\min \limits_{\tiny {\pi\in \prod \!N_{\tiny{d}}}} \!\!\!\sum_{\!\!\tiny{i=0}}^{\tiny {N_{d}-1}}\! \!\left\{\!\alpha L_{{lane}}\left(\hat{v}_{\pi(i)}, \!v_{i}\right),\!\ \! \beta L_{{point}}\left(\hat{p}_{\pi(i)}, \!p_{i}\right)\!\right\}\!\!,\!\!
\end{equation}
\begin{equation}
L_{{lane}}=-\sum_{\tiny{i=0}}^{\tiny{N_{gt}-1}} \log_{}{\left ( \hat{v}_{i} \right ) } \left [ v_{i} \right ],
\end{equation}
\begin{equation}
L_{{point}}=\frac{1}{N_{gt}} {\sum_{\tiny{i=0}}^{\tiny N_{gt}-1}\sum_{\tiny{j=0}}^{\tiny {\frac{N_p}{2}}-1} \left\{\left| \hat{p}_{ij}^{l}-p_{ij}^{l}\right| + \left| \hat{p}_{ij}^{r}-p_{ij}^{r}\right|\right\}},
\end{equation}
where $L_{lane}$ and $L_{point}$ represent lane and point level alignment, and $N_{gt}$ indicates successful matches with ground truth, and in training, $\alpha$ and $\beta$ are in a 5:2 ratio.

\noindent\textbf{Double-Edge Prediction Loss. } \textcolor{mod_color}{Following the optimal alignment result, we train for perception and planning to predict double-edges. These include (1) $L_{{edge\_bev}}$ for BEV regression, (2) $L_{{int}}$ and (3) $L_{{dir}}$ for intersection and direction respectively at the lane level, and (4) $L_{{occ}}$ and (5) $L_{{plan}}$ for occupancy and planning respectively at the point level. Additionally, $L_{{speed}}$ and $L_{{signal}}$ are used for predicting planning speed and traffic signals. Formulated as:}
\begin{equation}
\begin{aligned}
Loss= \gamma L_{edge\_bev} + \delta L_{int} + \epsilon L_{dir} + \varepsilon L_{occ} +\\
\zeta L_{plan} + \eta L_{speed} +\theta L_{signal},
\end{aligned}
\end{equation}
\textcolor{mod_color}{where, in training, ${\gamma}$, ${\delta }$, ${\epsilon}$, ${\varepsilon}$, ${\zeta}$, ${\eta}$ and ${\theta}$are set to 5:2:1:3:4:1:0.1. $L_{{edge\_bev}}$ and $L_{{plan}}$ can be formulated as:}
\begin{equation}
\!L_{{edge\_bev}}\!=\!\frac{1}{N_{gt}}\!\! \sum_{\tiny {i=0}}^{\tiny {N_{gt}-1}}\!\sum_{\tiny {j=0}}^{\tiny {{\frac{N_p}{2}}-1}}\!\! \left \{ \left | pred_{ij}^{{l}}\!-\!gt_{ij}^{{l}} \right | \!+\! \left | pred_{ij}^{{r}} \!-\! gt_{ij}^{{r}} \right | \!\right \},
\end{equation}
\textcolor{mod_color}{
\begin{equation}
L_{plan}\!\!=\!\!\!\!\!\!\sum_{\tiny {i=0}}^{\tiny {N_{gt}-1}}\! \sum_{\tiny {j=0}}^{\tiny {{\frac{N_p}{2}}-1}}\! \!\!\left \{\!\! \frac{{\left ( \!\rho\!\cdot \!\left (\!1\!\!-\!e^{\tiny{\!-CE(pred_{ij}},gt_{ij})}\!   \right ) \!\right )^{\tiny {2}}\! \!\!\cdot\! CE(pred_{ij},gt_{ij})}}{D_{ij}^{p2t}} \!\!\right \}\!\!,\!
\end{equation}
}
\textcolor{mod_color}{where $D_{{p2t}}$ represents the distance from an edge point into the target vector, serving as weights to emphasize planning features near target points, $\mathit{CE}$ refers to the cross-entropy loss, and $\rho$ is set to 0.25. Additionally, $L_{{int}}$, $L_{{dir}}$, and $L_{{occ}}$ employ Focal Loss~\cite{tool-focal}, while $L_{{speed}}$ uses SmoothL1Loss\cite{tool-fastrcnn-smooth-l1}, and $L_{{signal}}$ is based on cross entropy loss. }
\section{Experiments}
\label{sec:experiments}
\subsection{Implementation Details}
\noindent\textbf{Dataset and Metrics.}
Using autonomous driving environment Carla\cite{carla}, we gather 126K frames from diverse scenarios across 8 maps and 13 weathers. The data are collected at 2Hz with vehicles equipped with four cameras, an IMU, and a GPS. We also annotate edge points with intersection, direction, occupancy, and planning attributes.
We follow the NEAT~\cite{m2m-neat} setup for both training and evaluation.
In Town05 benchmark, we trained on Town01$\sim$04, 06$\sim$07, and 10, and evaluated on Town05. In CARLA 42 Routes, we used all route dataset for training, like NEAT, and it's noteworthy that the evaluation routes differ from training.
Driving Score (DS), Route Completion (RC), and Infraction Score (IS) are key metrics evaluating performance, where higher scores reflect better progress, safety, and rule adherence.

\begin{figure*}[htp]
  \centering
  \includegraphics[width=\linewidth]{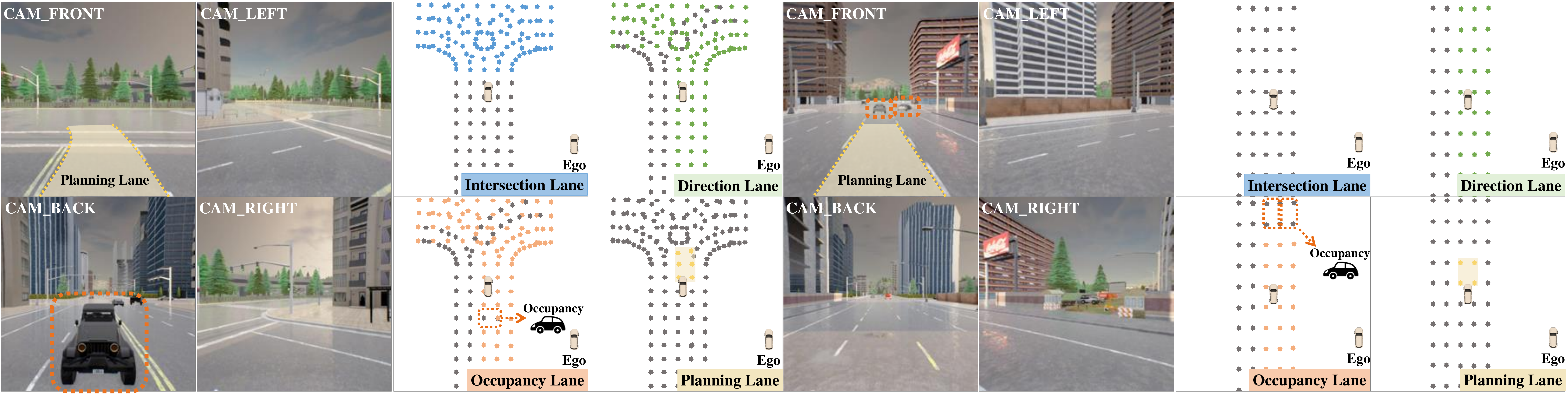}
   \caption{\textcolor{mod_color}{Visualization of \systemname includes multi-camera images capturing views from the front, back, left, and right, along with double-edges that detail intersection lanes, direction lanes, occupancy lanes, and selected lanes for planning. Within the occupancy lane visualization, lanes occupied by vehicles are marked with orange dashed rectangles. Planning lanes are highlighted in yellow on the front camera, enhancing the visibility of the intended lane for the vehicle.} 
   }
  \label{fig:exp_vis}
  \vspace{-5pt}
\end{figure*}

\noindent\textbf{Component Details.}
We scale $800\times600$ images to $224\times224$ and process them using ResNet-50, with settings: ${{N_{d}}=30}$, ${{N_{p}}=20}$, 6 layers each for the encoder and decoder.
In the hierarchical feature fusion, features are fused and fed into the target-guided planning branch, where the target vector and planning embeddings are processed with an attention decoder (feature dimension: 256, hidden layer: 128).

\begin{table}[t]
\centering
\caption{Performance Comparison on Carla Town05.}
\setlength\tabcolsep{8.0pt}
    \begin{tabular}{l|cc|cc}
    \toprule
        & \multicolumn{2}{c|}{\textbf{Town05\  Short}} & \multicolumn{2}{c}{\textbf{Town05\  Long}} \\
        \hline
        \textbf{Method}& \makecell{DS $\uparrow$ (\%)}& \makecell{RC $\uparrow$ (\%)}& \makecell{DS $\uparrow$ (\%)}  & \makecell{RC $\uparrow$ (\%)} \\
        \hline
        CILRS\cite{e2e-cilrs}& 7.47& 13.40& 3.68& 7.19\\
        LBC\cite{e2e-lbc}& 30.97& 55.01& 7.05& 32.09\\
     ST-P3\cite{m2m-stp3}& 55.14& 86.74& 11.45&\textcolor{gray}{\textbf{83.15}}\\
     VAD\cite{m2m-vad}& 64.29& 87.26& 30.31&75.20\\
     NEAT\cite{m2m-neat}& 58.70& 77.32& 37.72&62.13\\
     WOR\cite{m2m-wor}& 64.79& 87.47& \textcolor{gray}{\textbf{44.80}}&82.41
    \\
     Roach\cite{m2m-roach}& \textcolor{gray}{\textbf{65.26}}& \textcolor{gray}{\textbf{88.24}}& 43.64&80.37\\
         \hline
     \textbf{\systemname \ ($ours$)}& \textbf{92.46}& \textbf{97.04}& \textbf{78.27}&\textbf{96.16}\\
        \hline
        \textbf{Impro. (\%)} & \textbf{+27.20}& \textbf{+8.80}& \textbf{+33.47}& \textbf{+13.01}\\
    \bottomrule
    \end{tabular}
\label{tab:carla-t5}
\vspace{-5pt}
\end{table}

\begin{table}[t]
\centering
\caption{Performance Comparison on Carla 42 Routes.}
    \setlength\tabcolsep{13.5pt}
    \begin{tabular}{l|ccc}
    \toprule
        & \multicolumn{3}{c}{\textbf{Carla \ 42 \ Routes}} \\
        \hline
        \textbf{Method}& \makecell{DS $\uparrow$ (\%)}& \makecell{RC $\uparrow$ (\%)}& \makecell{ IS $\uparrow$ (\%)}\\
        \hline
        CILRS\cite{e2e-cilrs}& 22.97& 35.46& 66.00\\
        LBC\cite{e2e-lbc}& 29.07& 61.35& 57.00\\
     AIM\cite{m2m-multi-moda-fusion}& 51.25& 70.04&73.00\\
     TransFuser\cite{m2m-transfuser} & 53.40& 72.18& 74.00\\
     NEAT\cite{m2m-neat}& 65.17& 79.17& \textcolor{gray}{\textbf{82.00}}\\
     Roach\cite{m2m-roach}& 65.08 & 85.16& 77.00\\
     WOR\cite{m2m-wor}& \textcolor{gray}{\textbf{67.64}}& \textcolor{gray}{\textbf{90.16}}& 75.00 \\
         \hline
     \textbf{\systemname \ ($ours$)}&  \textbf{83.18}&  \textbf{94.02}& \textbf{89.00}  \\
        \hline
        \textbf{Impro. (\%)} & \textbf{+15.54}& \textbf{+3.86}& \textbf{+7.00}\\
    \bottomrule
    \end{tabular}
\label{tab:carla-42r}
\vspace{-15pt}
\end{table}

\subsection{Performance on Carla Benchmark}

\begin{table*}[t]
\centering
\caption{Ablation study on Carla Town01, Town05 and 42 Routes. ``TGP" denotes target-guided planning,  ``HEF" denotes hierarchical feature early fusion, and ``DLF" denotes the double-edge late-fusion enabled in the double-edge interpreter.}
\setlength\tabcolsep{6.0pt}
    \centering
        \begin{tabular}{c|ccc|ccc|ccc|ccc|ccc} 
        \toprule
                \textbf{ID.}&  \textbf{TGP}&  \textbf{HEF}&  \textbf{DLF}&\multicolumn{3}{c|}{\textbf{Carla Town01}}&\multicolumn{3}{c|}{\textbf{Town05 Short}}&\multicolumn{3}{c|}{\textbf{Town05 Long}}&\multicolumn{3}{c}{\textbf{Carla 42 Routes}}\\
                \hline
                \textbf{\%,$\uparrow$}&Sec\ \ref{method:drive-lane-prediction}&Sec\ \ref{method:dual-lane-fusion}&Sec\ \ref{method:dual-lane-interpreter}&  \textbf{RC}&  \textbf{DS}&  \textbf{IS}&  \textbf{RC}&  \textbf{DS}&  \textbf{IS}&  \textbf{RC}&  \textbf{DS}&  \textbf{IS}&  \textbf{RC}&  \textbf{DS}&  \textbf{IS}\\ 
             \hline
        A&\tiny{\color{red}\XSolidBrush}&\tiny{\color{red}\XSolidBrush}&\tiny{\color{red}\XSolidBrush}& 25.09& 19.86& 72.00& 27.05& 23.82& 88.80& 10.86& 7.99& 58.90& 32.41& 14.63&49.40\\
        B&\tiny{\color{green}\CheckmarkBold}&\tiny{\color{red}\XSolidBrush}&\tiny{\color{red}\XSolidBrush}& 86.89& 47.02& 51.90& 81.85& 54.85& 67.70& 74.96& 22.99& 26.20& 86.12& 59.99&68.70\\
        C&\tiny{\color{green}\CheckmarkBold}&\tiny{\color{green}\CheckmarkBold}&\tiny{\color{red}\XSolidBrush}& 88.57& 75.64& 83.10& 90.17& 75.85& 84.10& 81.86& 43.63& 51.50& 90.74& 70.35&76.90\\ 
         \hline
        \textbf{D}&\tiny{\color{green}\CheckmarkBold}&\tiny{\color{green}\CheckmarkBold}&\tiny{\color{green}\CheckmarkBold}&  \textbf{90.20}&  \textbf{87.48}&  \textbf{96.00}&  \textbf{97.04}&  \textbf{92.46}&  \textbf{95.10}&  \textbf{96.16}&  \textbf{78.27}&\textbf{81.30}&  
        \textbf{94.02}&  \textbf{83.18}&\textbf{89.00}\\
         \hline
        \end{tabular}
      \vspace{-15pt}
    \label{tab:ablation}
\end{table*}
In this section, we evaluate \systemname's performance across three benchmarks: Town05 long, Town05 short, and Carla 42 Routes, which challenge autonomous driving algorithms with diverse urban and rural settings, dynamic objects, and varying weather. The benchmarks range from complex urban environments to village roads, including features like highways and roundabouts. Comparative results with state-of-the-art \textbf{vision-based algorithms} are presented in Tables \ref{tab:carla-t5} and \ref{tab:carla-42r}.

\noindent\textbf{Driving Score.}
The Driving Score (DS) comprehensively assesses autonomous driving systems performance, combining route completion and infraction scores. In the Town05 Long, Short, and Carla 42 Routes benchmarks, \systemname excels over other algorithms, achieving the highest DS scores. Specifically, in the Town05 benchmarks, our system achieves a DS improvement of $27.20\%$ (Short) and $33.47\%$ (Long), as detailed in Table \ref{tab:carla-t5}. Additionally, in the Carla 42 Routes Benchmark, a notable $15.54\%$ increase in DS score is observed, as in Table \ref{tab:carla-42r}. These results underscore the effectiveness of our approach in enhancing planning safety and traffic regulation compliance.

\noindent\textbf{Route Completion.}
Route Completion (RC) is a critical metric for assessing an autonomous driving system's success in completing predetermined routes throughout the evaluation process, reflecting the system's ability to plan routes and accurately understand target points. In the Carla Town05 and Carla 42 Routes benchmark, our method, by integrating planning with perception, improves its understanding of the traffic environment and target points, demonstrating exceptional route completion capabilities and achieving the highest scores. Compared to other algorithms, our method shows a higher route completion rate in benchmark tests, including a diverse range of urban and rural environments. Specifically, on the Carla Town05 Short and Long benchmarks, our method achieves improvements of $8.80\%$ and $13.01\%$, respectively, as shown in Table \ref{tab:carla-t5}. Additionally, on the Carla 42 Routes benchmark, which features a variety of scenarios, our method also achieves an improvement of $3.86\%$, as indicated in Table \ref{tab:carla-42r}. These results highlight our method's advanced capabilities in route planning and target point understanding and its robustness in adapting to different road conditions.

\noindent\textbf{Infraction Score.}
Infraction Score (IS) is a comprehensive metric used to evaluate systems performance, including avoiding collisions, adhering to traffic rules, and handling complex situations. 
\systemname exhibits significant performance in Carla 42 Routes, achieving a $7.00\%$ enhancement in Infraction Score (IS) as detailed in Table \ref{tab:carla-42r}, surpassing other algorithms.

\subsection{Qualitative Results}
In Figure \ref{fig:exp_vis}, we illustrate our system's (\systemname) ability to hierarchically perceive the traffic environment and seamlessly integrate planning tasks at the lane level. The visualization includes intersection lanes marked in blue, direction lanes that indicate roads complying with traffic regulations marked in green, and occupancy lanes for roads unoccupied by traffic agents and adhering to directions marked in orange. The planning lane, highlighted in yellow, signifies the optimally chosen lane that ensures safety and leads to the target point. In addition, we include camera images from four perspectives. 

\subsection{Ablation Studies}
\noindent\textbf{Impact of Target-Guided Planning Branch: }
In \systemname, analyzing the correlation between traffic environment information and target points can significantly enhance decision accuracy and routing completion. The target-guided planning branch (TGP) deepens understanding of this correlation, optimizing decision-making. The TGP identifies key planning features related to the target vector, strengthening their decision-making role.
Experiment B, using TGP, the RC metric shows increases of ${61.80\%}$, ${54.80\%}$, ${64.10\%}$, and ${53.71\%}$ over Experiment A, as shown in Table \ref{tab:ablation}.
This underscores the TGP module's effectiveness in guiding decisions and the value of understanding the correlation between the target vector and planning features. Aligning decisions with the target vector improves accuracy and route completion.

\noindent\textbf{Impact of Hierarchical Feature Early Fusion Module: }
Although the introduction of the TGP in \systemname improves route completion, it also highlights a limitation: the planning features' inability to fully comprehend the environment, leading to a reduction in infraction score. 
The inclusion of the TGP and the Hierarchical Early Fusion (HEF) module in Experiment C led to noticeable improvements in both infraction and driving scores. When compared with Experiment B, the infraction scores increased by ${31.20\%}$, ${16.40\%}$, ${25.30\%}$, and ${8.20\%}$. The driving scores also increased by ${28.62\%}$, ${21.00\%}$, ${20.64\%}$, and ${10.36\%}$, as shown in Table \ref{tab:ablation}.
This indicates that optimizing planning through TGP alone is insufficient; it is necessary to enhance environmental understanding through modules like HEF to achieve a deep integration of planning and perception. 
The core of HEF lies in its ability to understand the environment hierarchically and explore the correlations between multiple layers of perception for attention fusion, significantly improving the environmental perception capabilities related to planning. 
Since planning tasks are constrained by perception, the safety of planning is further enhanced. This underscores the importance of deeply integrating planning and perception.

\noindent\textbf{Impact of Double-Edge Interpreter: }
In \systemname, the path generation interprets road geometry, traffic, and planning attributes, converting them into planning paths. This process is indispensable. The DLF, an additional late-fusion module, enhances safety by leveraging perception and planning information. It late-fuses this information to avoid hazardous planning, thus improving safety. 
Experiment D validates the effectiveness of DLF in coordinating perception and planning, with infraction score increases of ${12.90\%}$, ${11.0\%}$, ${29.80\%}$, and ${12.10\%}$ compared to Experiment C, which lacked DLF.
This strategy extracts safety-related information from an intersection, direction, occupancy, and planning attributes, significantly lowering infraction rates and improving driving performance.

\noindent\textbf{Efficiency of \systemname: }
\begin{table}[t]
	\centering
    \caption{\textcolor{mod_color}{A comparative analysis of latency and frames per second (FPS) across methods using multi-view camera inputs, where lower latency correlates with higher FPS.}}
    \setlength\tabcolsep{20.0pt}
	\begin{tabular}{c|c|c|c}
        \toprule
        \multicolumn{2}{c|}{\textbf{Method}} & \textbf{Latency (ms)  $\downarrow$} & \textbf{FPS  $\uparrow$}  \\
        \hline
        \multicolumn{2}{c|}{ST-P3\cite{m2m-stp3}} & 476.74 & 2.1  \\
        \multicolumn{2}{c|}{TransFuser\cite{m2m-transfuser}} & 171.93 & 5.82  \\
        \multicolumn{2}{c|}{NEAT\cite{m2m-neat}} & 85.08 & 11.75  \\
        \multicolumn{2}{c|}{VAD\cite{m2m-vad}} & 59.50 & 16.81  \\
        \hline
        \multicolumn{2}{c|}{\textbf{\systemname (ours)}} & \textbf{44.30} & \textbf{22.57}  \\
        \bottomrule
	\end{tabular}
      \vspace{-15pt}
    \label{tab:efficiency}
\end{table}
At the core of \systemname's innovation is the design of the lane-level data structure, which confines planning and perception tasks to the lane space, significantly reducing the computational cost. 
The effectiveness of this integrated approach is underscored by a significant decrease in total inference time to 44.30 ms and an impressive frame rate of approximately 22.57 FPS, as shown in Table \ref{tab:efficiency}.

\section{Conclusion}
\label{sec:conclusion}
In this paper, we propose Perception Helps Planning (\systemname), a novel framework that innovatively integrates planning with perception tasks at the lane level. Distinct from previous efforts, \systemname employs a specific data structure to transform path planning into a lane-level task, embedding the planning process deeply within perception tasks. This unique strategy not only enables simultaneous traffic perception and planning within lanes but also ensures that the planning approaches are thoroughly compliant with traffic regulations.
Moreover, this integration significantly improves efficiency and ensures adherence to traffic regulations, thereby enhancing safety and reliability.
Experiments on Carla Benchmark demonstrate the effectiveness of the proposed method, which outperforms the state-of-the-art vision-based end-to-end planning algorithms.

\bibliographystyle{IEEEtran}
\bibliography{IEEEabrv, 7-reference}

\end{document}